\documentclass[conference]{IEEEtran}
\IEEEoverridecommandlockouts
\usepackage{cite}
\usepackage{amsmath,amssymb,amsfonts}
\usepackage{algorithmic}
\usepackage{graphicx}
\usepackage{textcomp}
\usepackage{xcolor}
\def\BibTeX{{\rm B\kern-.05em{\sc i\kern-.025em b}\kern-.08em
    T\kern-.1667em\lower.7ex\hbox{E}\kern-.125emX}}

\newcommand \ignore[1]{}
\newcommand*{\affaddr}[1]{#1} 
\newcommand*{\affmark}[1][*]{\textsuperscript{#1}}
\newcommand*{\email}[1]{\texttt{#1}}
\usepackage{url}
\usepackage{colortbl}

\usepackage{caption}
\usepackage{subcaption}

\makeatletter
\def\ps@IEEEtitlepagestyle{%
  \def\@oddfoot{\mycopyrightnotice}%
  \def\@oddhead{\hbox{}\@IEEEheaderstyle\leftmark\hfil\thepage}\relax
  \def\@evenhead{\@IEEEheaderstyle\thepage\hfil\leftmark\hbox{}}\relax
  \def\@evenfoot{}%
}
\def\mycopyrightnotice{%
  \begin{minipage}{\textwidth}
  \scriptsize
  ~\copyright~2024 IEEE.  Personal use of this material is permitted.  Permission from IEEE must be obtained for all other uses, in any current or future media, including reprinting/republishing this material for advertising or promotional purposes, creating new collective works, for resale or redistribution to servers or lists, or reuse of any copyrighted component of this work in other works.
  \end{minipage}
}
\makeatother

\begin{document}

\title{
Toward Mitigating Sex Bias in Pilot Trainees' Stress and Fatigue Modeling
}

\author{Rachel Pfeifer\affmark[1], Sudip Vhaduri\affmark[2], Mark Wilson\affmark[3], and Julius Keller\affmark[4]\\
\affaddr{\affmark[1]Computer Science Department,
\affmark[2]Computer and Information Technology Department,\\
\affmark[3]School of Health Sciences,
\affmark[4]School of Aviation and Transportation Technology, Purdue University, IN 47907}\\
\email{\{\affmark[1]rjpfeife,\affmark[2]svhaduri,\affmark[3]wilso774,\affmark[4]keller64\}@purdue.edu}
}

\maketitle

\begin{abstract}
While researchers have been trying to understand the stress and fatigue among pilots, especially pilot trainees, and to develop stress/fatigue models to automate the process of detecting stress/fatigue, they often do not consider biases such as sex in those models. However, in a critical profession like aviation, where the demographic distribution is disproportionately skewed to one sex, it is urgent to mitigate biases for fair and safe model predictions.  
In this work, we investigate the perceived stress/fatigue of 69 college students, including 40 pilot trainees with around 63\% male. We construct models with decision trees first without bias mitigation and then with bias mitigation using a threshold optimizer with demographic parity and equalized odds constraints 30 times with random instances.
Using bias mitigation, we achieve improvements of 88.31\% (demographic parity difference) and 54.26\% (equalized odds difference), which are also found to be statistically significant. 
\end{abstract}


\section{Introduction}\label{introduction}

We first introduce the motivation and importance of this work, followed by related work in this area and our contribution in this section.

\subsection{Motivation}
Flying planes are considered one of the most stressful jobs, and pilots often suffer from fatigue and their adverse impacts on health and well-being. Recent studies found that 40\% of pilots feel the most stress from the number of flights they have, and over 50\% of pilots are stressed due to fear of losing their job \cite{pilotstatistics}. Additionally, approximately 43\% of pilots do not recommend their career field to young people \cite{pilotstatistics}. These amounts of stress are also significant for students in professional flight programs when compared to their peers in other academic areas, as they undergo a high level of training and learning to become pilots.

It is important to implement machine learning models that predict whether a pilot is feeling stressed or fatigued to automate the process. These models can reveal what contributes to elevated stress and fatigue levels in pilots and how different factors can influence these levels. But, there is an overwhelming amount of sex bias in the industry. As of 2023, the International Civil Aviation Organization reports that only 4\% of pilots are female, the rest being male \cite{pilotgenderstatistics}. Such bias can compromise the model's effectiveness. Therefore, there is a critical need to take into account the demographic bias, such as sex bias, that is present in the aviation industry when developing machine learning models with pilots' data. 

\subsection{Related Work}
Studies have found significant stress and fatigue levels experienced by pilots in the aviation industry. A team of researchers discusses the particular risks associated with high levels of stress and the safety concerns that these risks pose~\cite{stressandfatigue}. Another study discusses the sources of aviation failure and how they are connected to the neurological and cognitive responses of pilots~\cite{neurologicalandcognitiveresponses
}. It is noted that the most identifiable and preventable sources of aviation failures are connected with fatigue. Since it is preventable, understanding the cause of this problem is important in improving flight safety~\cite{preventingfatigue}. Students in the aviation industry dedicate a great amount of time to preparing and completing aviation training, adding to the levels of fatigue that they already experience. Existing studies reflect this by analyzing the stress levels of professional pilots who underwent this rigorous training. These studies employ surveys to quantify fatigue indicators, such as the multidimensional fatigue inventory (MFI)~\cite{pilotnonpilot}. 

Current research with the presence of demographic bias in datasets, specifically in the healthcare industry, is causing trouble and thereby getting attention in recent times. Some of these works focus on reducing statistical and algorithmic bias in AI models that are being developed to diagnose medical conditions \cite{statisticalandalgorithmicbias}. Bias in AI models can lead to incorrect diagnoses or surgical outcomes, depending on how the AI model is used. Underrepresented groups are negatively impacted in terms of what and how much medical care they receive, especially since these AI models amplify preexisting bias~\cite{aibias}. Additional research involves reducing racial bias in clinical machine learning models to make the algorithms fair~\cite{racialbias}. Yet, stress and fatigue models developed with pilot data do not address mitigating sex bias~\cite{aifatiguemodels}. 

\subsection{Contributions}
The main contribution of this work is to determine how machine learning models developed with stress and fatigue data obtained from pilot trainees are impacted by demographic bias, specifically sex bias (male versus female), as pilot sex distribution is significantly skewed to one sex, i.e., there are a greater number of male pilots than female pilots in the aviation industry. By applying two major bias mitigation algorithms (i.e., threshold optimizer with demographic parity and equalized odds constraints), we are able to reduce the demographic bias significantly across males and females.

\section{Methodology}\label{methods}

In this section, we will present our methodology to mitigate bias in pilot/non-pilot fatigue models.
However, before presenting the bias mitigation methodology and dataset with pre-processing steps, we will first define the key terminologies.

\subsection{Preliminaries}\label{preliminaries}

\subsubsection{Fairness and Bias}
According to the FairLearn library~\cite{fairlearn} used in this work, fairness is defined in terms of the unfair behavior or the harm a system, specifically an AI system or model, has on groups of people. Bias is defined as the types of harm that are inflicted on groups of people. In this paper, we focus on demographic bias, specifically sex bias, i.e., the unfair impact caused by models or AI systems due to sex variation.

\subsubsection{Fairness Metrics}\label{fairMetrics}
We use the following six metrics when assessing the performance of our bias mitigation algorithms.

\begin{itemize}
    \item Selection rate is defined as the ratio of predicted labels that match the correct outcome. Because of the implications of this, it is essential to maximize this metric when we mitigate the bias in a dataset.
    \item Demographic parity ratio is the fraction of the smallest group selection rate and the greatest group selection rate. Bias mitigation should maximize the demographic parity ratio metric.
    \item Demographic parity difference is the difference between the smallest group selection rate and the greatest group selection rate. Bias mitigation should minimize the demographic parity difference metric.
    \item False negative rate is defined as the probability that a machine learning model predicts a negative value when the actual value is positive. This is also known as a type II error, and in terms of statistical analyses, means not rejecting the null hypothesis when it is false. Because of these implications, minimizing the false negative rate when we mitigate the bias in a dataset is essential.
    \item Equalized odds ratio is defined as the minimum of the true positive rate and the false positive rate. If the equalized odds ratio is equal to 1, this means that the dataset has the same false negative rate, true positive rate, false positive rate, and true negative rate. Bias mitigation should maximize the equalized odds ratio. 
    \item Equalized odds difference is the maximum of the true positive rate difference and the false positive rate difference. The true positive rate difference and false positive rate difference are similar to the true positive rate and false positive rate defined above. However, with the equalized odds difference, we are calculating the difference in these rates. Bias mitigation should minimize the equalized odds difference metric.
\end{itemize}

\subsubsection{Sensitive Feature and Constraint}
A sensitive feature refers to an attribute that pertains to an individual's identity, e.g., sex, age, and ethnicity. 
A constraint refers to criteria applied to the bias mitigation algorithms to affect their impact on the sensitive feature. The mitigation algorithms match fairness metrics across the sensitive feature by adhering to these constraints, e.g., demographic parity or equalized odds. 

\subsubsection{Percentage Improvement}
To compare the performance of our bias mitigation algorithms, we utilize percent improvement, which is defined as the absolute value of the difference in the value of a metric prior to and following mitigation divided by its value before mitigation and then multiplied by 100.  

\begin{figure*}
\centering
\subfloat[Selection rate]{\includegraphics[width=.31\linewidth]{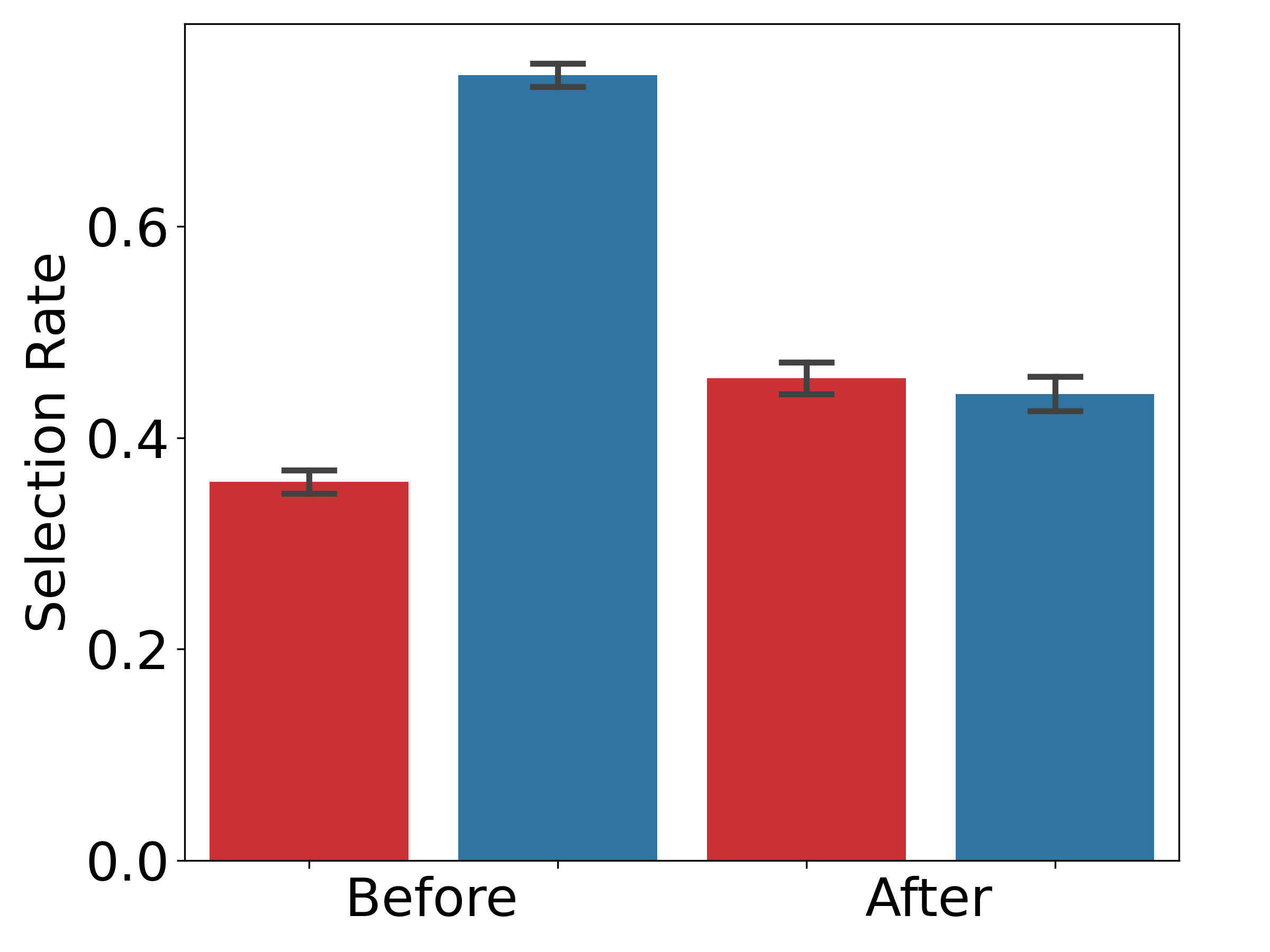}\label{selectionrate}}
\hfill
\subfloat[Demographic parity ratio]{\includegraphics[width=.31\linewidth]{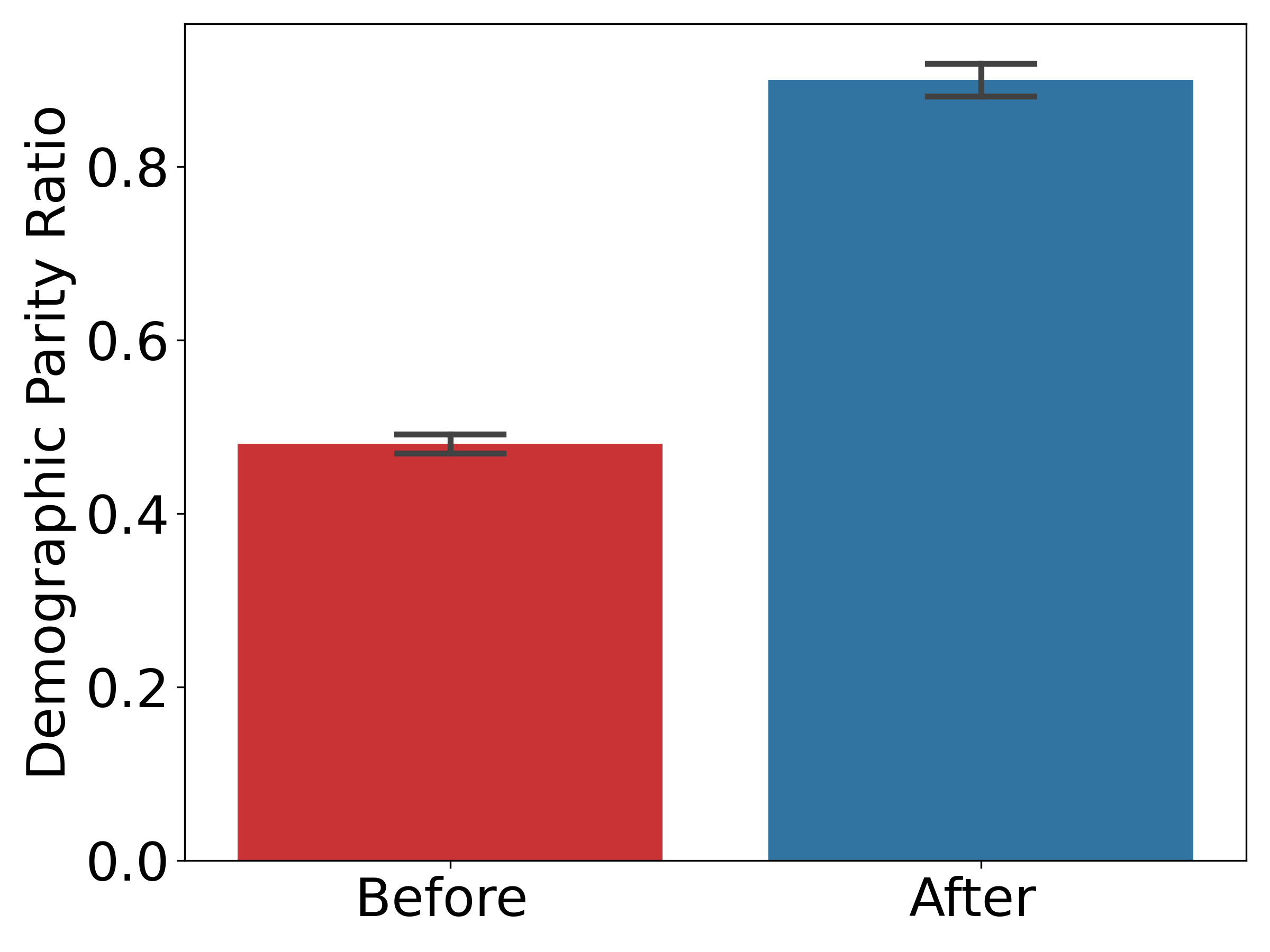}\label{demographicparityratio}}
\hfill
\subfloat[Demographic parity difference]{\includegraphics[width=.31\linewidth]{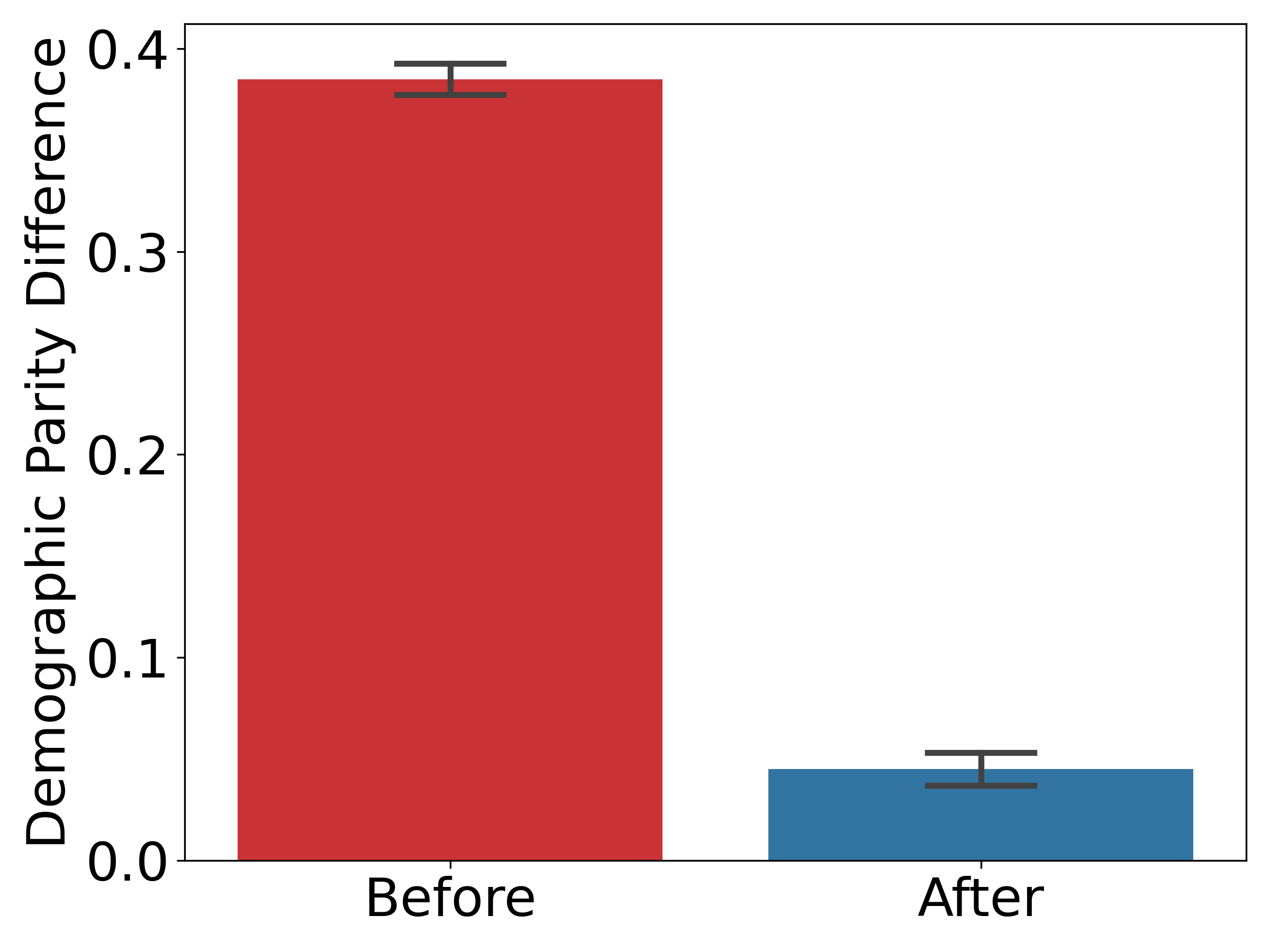}\label{demographicparitydifference}}
\caption
{Assessment of demographic parity constraint [in sub-figure (a) red and blue bars representing female and male].}
\label{demographicparity}
\end{figure*}

\subsection{Data Collection}\label{dataCollect}
Our dataset comes from the survey results of college students in professional flight programs (IRB-2023-12) and other academic programs (IRB-2022-1383) at Purdue University. We identify this list of students through the Purdue Institutional Data Analytics + Assessment (IDA + A) Center. 
Our collected dataset includes participants' sex, age, as well as responses to the perceived stress scale (PSS), Jenkins sleep scale (JSS), multidimensional fatigue inventory (MFI), general fatigue (GF), physical fatigue (PF), reduced activity (RA), reduced motivation (RM), and mental fatigue (MF). Our initial dataset contains responses from 40 pilot students (15 are female and 25 are male) and 29 non-pilot students (22 are female and seven are male)~\cite{pilotnonpilot}.

\subsection{Pre-processing}\label{preProc}
We first drop participants who completely skipped a set of questions used to calculate the stress/fatigue scores. Next, we perform scale inversion for some of the questions (as needed to match polarity~\cite{pilotnonpilot}) and aggregation to get a stress or fatigue score. While aggregating individual responses to get a score, we found some sets were responded partially. Therefore, we calculate an aggregated normalized score, $s = \frac{\sum_{i = 1}^{n} r_i}{n \times m}$, where $n$, $r_i$, and $m$ are the total number of questions in a set, response to the $i^{th}$ question in the set, and the maximum numeric value of response, with the normalized aggregated score $s$ with $0 \le s \le 1$. After all the processing, we have 28 pilot students (9 female) and 20 non-pilot students (14 female).

\subsection{Data Modeling}
Our feature set comprises sex (the sensitive feature analyzed in this manuscript), age, PSS, JSS, MFI, GF, PF, RA, RM, and MF with pilot/non-pilot class labels. We choose to use decision tree classification learners. For class balancing, we randomly pick 20 pilot instances whenever developing a decision tree model.

\subsection{Hyper-parameter Optimization}
We utilize grid search cross-validation ($CV$) to find the optimal values of the hyper-parameters used for the decision tree model. In this work, we optimize four types of parameters: the criterion (with ``gini'' or ``entropy'' options), $max\_depth$ (with values $[3,15]$), $min\_samples\_leaf$ (with values $\{2,3,4,5\}$), and $min\_samples\_split$ (with values $\{2,3,4,5\}$). We find $CV = 7$ as an optimal choice while making a trade-off between bias and variance, with ``gini,'' $max\_depth = 3$, $min\_samples\_leaf = 2$, and $min\_samples\_split = 2$ as optimal hyper-parameter values. 

\subsection{Bias Mitigation}
We employ the FairLearn Python library~\cite{fairlearn} to mitigate bias in datasets and models trained with sex as a sensitive feature using two algorithms, i.e., threshold optimizer with demographic parity and equalized odds constraints with three metrics (defined in Section~\ref{fairMetrics}). 
In FairLearn, a threshold optimizer is a post-processing method and a classifier we use by applying a unique threshold for each group to the estimator. In our dataset, the group we have is sex (male and female), and the estimator we use is a decision tree. Simply put, we are applying a threshold to sex in our decision tree when we utilize the threshold optimizer. 
We calculate the value of each metric 30 times before and after the mitigation algorithms are applied. Our codes can be found in GitHub~\cite{pilotgithub}.

\begin{figure*}
\centering
\subfloat[False negative rate]{\includegraphics[width=.31\linewidth]{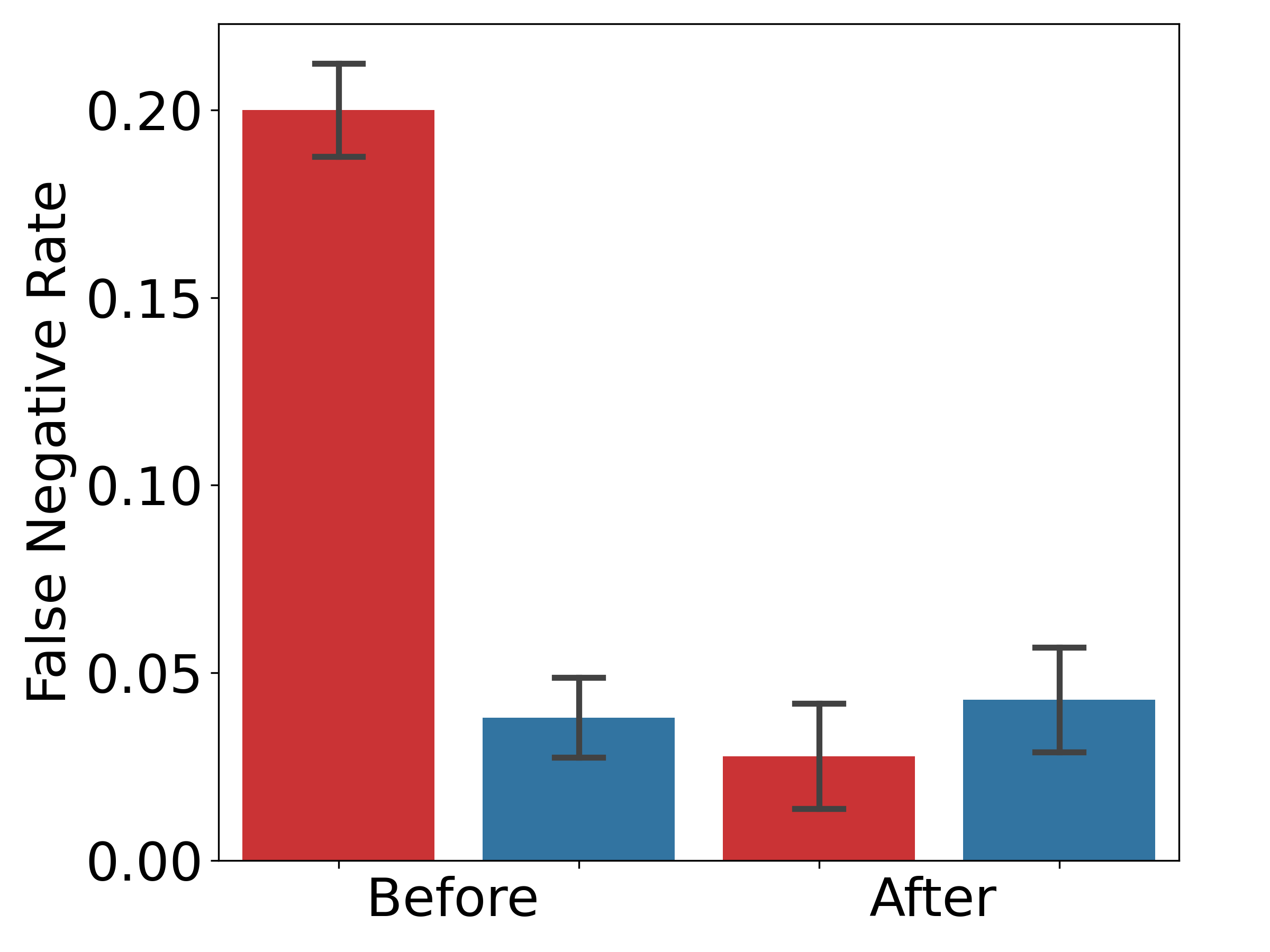}\label{falsenegativerate}}
\hfill
\subfloat[Equalized odds ratio]{\includegraphics[width=.31\linewidth]{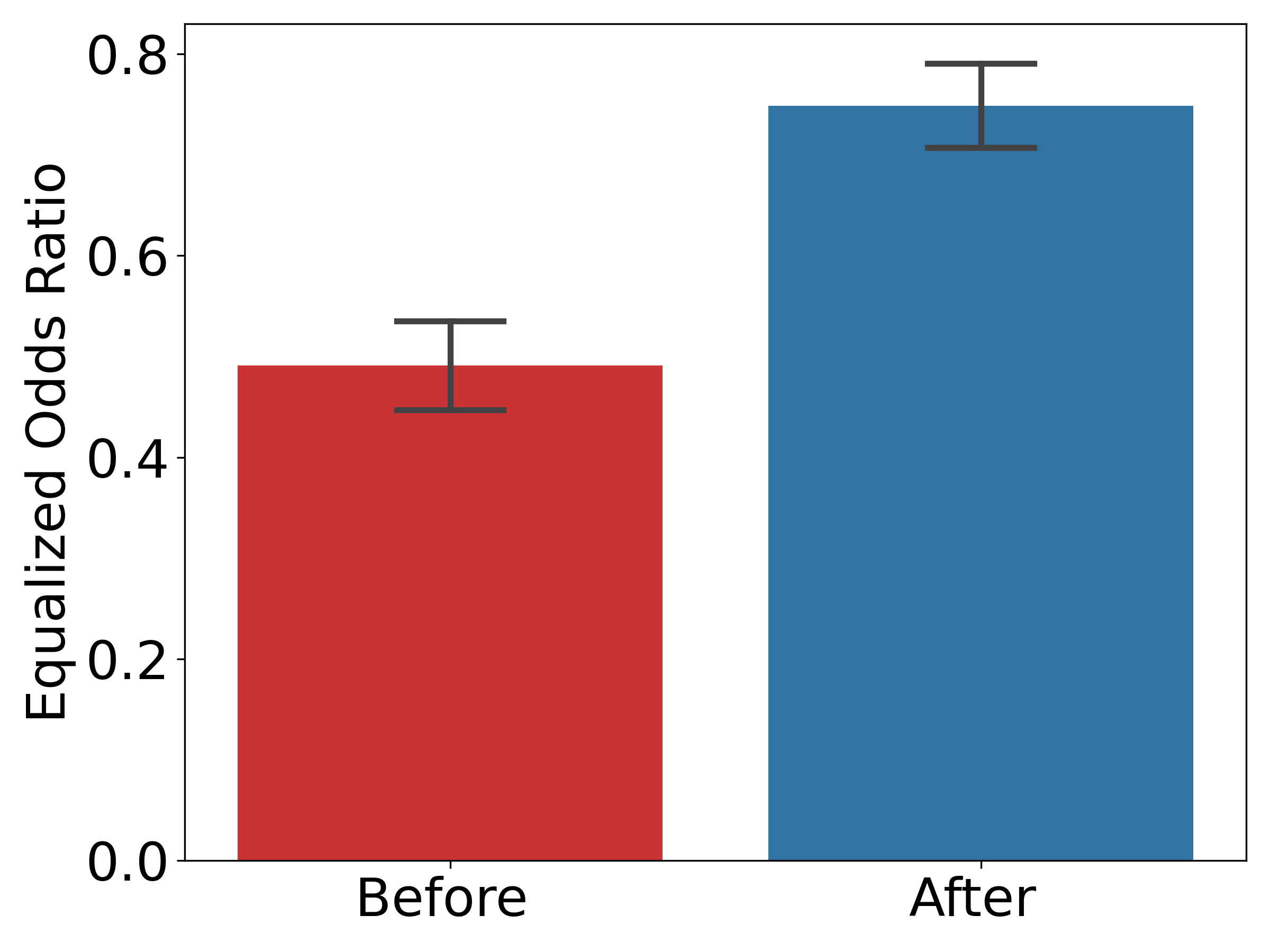}\label{equalizedoddsratio}}
\hfill
\subfloat[Equalized odds difference]{\includegraphics[width=.31\linewidth]{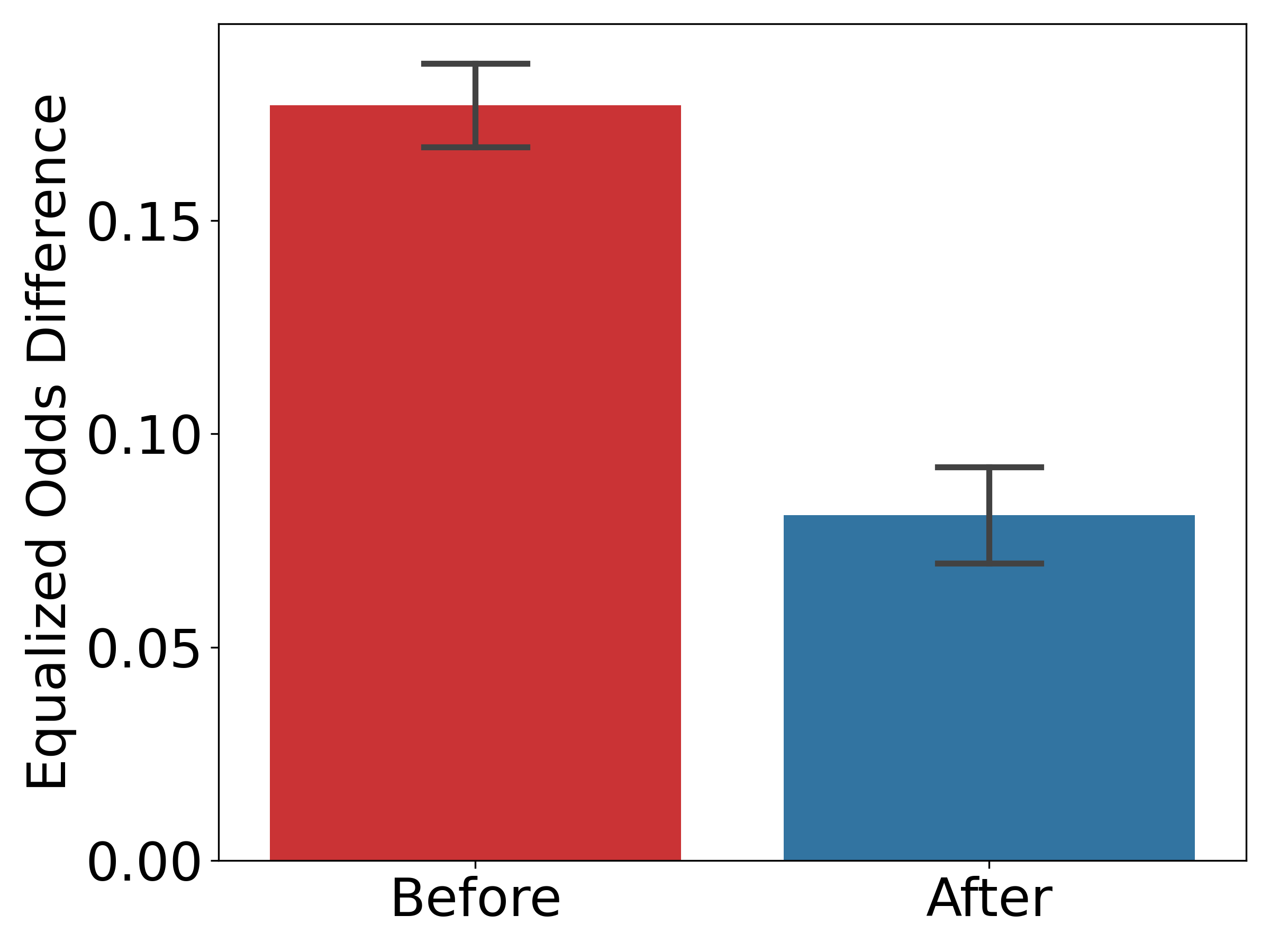}\label{equalizedoddsdifference}}
\caption{Assessment of equalized odds constraint [in sub-figure (a) red and blue bars representing female and male].}
\label{equalizedodds}
\end{figure*}

\section{Analysis}\label{analysis}

We analyze the performance of our bias mitigation modeling using graphical presentation and statistical tests. We use the ``percentage improvement'' to determine how much each metric changes after applying our mitigation algorithms (i.e., the threshold optimizer with demographic parity or equalized odds constraints). First, we analyze the demographic parity constraint (Section~\ref{dpAnalysis}), followed by the equalized odds constraint (Section~\ref{eoAnalysis}).

\subsection{Analysis of Demographic Parity Constraint}\label{dpAnalysis}
To assess the effect of bias mitigation on the demographic parity constraint, we use three common metrics, i.e., selection rate (to obtain equal selection rates across males and females), demographic parity ratio, and demographic parity difference. 

\subsubsection{Selection Rate}
In Figure~\ref{selectionrate}, we observe the female and male selection rates before and after mitigation. The female and male selection rates before mitigation are 35.83\% and 74.33\%, respectively. After mitigation, the female selection rate increases to 45.67\%, and the male selection rate decreases to 44.17\%, resulting in a more balanced selection rate than before mitigation. The threshold optimizer algorithm focuses on fairness constraints even at the expense of certain performance metrics such as accuracy. Further, the demographic parity fairness constraint tries to equalize the selection rates between groups, which may decrease the average selection rate after mitigation in some groups, e.g., females in this case.

\subsubsection{Demographic Parity Ratio}
The demographic parity ratios before and after mitigation are shown in Figure~\ref{demographicparityratio}. We find that the demographic parity ratio is 48.03\% before mitigation, and it increases to 89.98\% after mitigation. Therefore, the bias mitigation algorithm improves the parity ratio by 87.34\%.

\subsubsection{Demographic Parity Difference}
Figure~\ref{demographicparitydifference} shows the demographic parity differences for the dataset. Before mitigation, the demographic parity difference is 38.50\% and decreases to 4.50\% after mitigation. This means that reducing the bias leads to an 88.31\% improvement in the parity difference metric. 

\subsubsection{Statistical Significance Test}
We perform a \emph{Student's Two-sample t-test} to determine if the demographic parity difference is statistically significant. Our null hypothesis is \emph{$H_0: \mu_b = \mu_a$}. Here, \emph{$\mu_b$} is the average demographic parity difference before mitigation, and \emph{$\mu_a$} is the average demographic parity difference after mitigation. The \emph{Student's Two-sample t-test} achieves \emph{$t(58) = 30.59$}, with \emph{p-value = $1.79e^{-37}$}. The \emph{p-value} is less than \emph{$\alpha = 0.05$}, so we can reject the previous null hypothesis. Thus, the results of the demographic parity difference are statistically significant.

\subsection{Analysis of Equalized Odds Constraint}\label{eoAnalysis}
To assess the effect of bias mitigation on equalized odds constraint, we use three common metrics: false negative rates (to achieve equal errors across males and females), the equalized odds ratio, and the equalized odds difference.

\subsubsection{False Negative Rate}
We present the female and male false negative rates before and after mitigation in Figure~\ref{falsenegativerate}. The female false negative rate is 20.00\% before mitigation and drops to 2.78\% after mitigation. The male false negative rate, however, increases slightly due to mitigation, going from 3.81\% to 4.29\%. We can see that after mitigation, the female and male false negative rates are more balanced and conclude that the mitigation algorithm decreases the average false negative rate, as noted by a decrease from 11.90\% to 3.53\%.

\subsubsection{Equalized Odds Ratio}
In Figure~\ref{equalizedoddsratio}, the equalized odds ratios for the dataset is illustrated. Before mitigation, the equalized odds ratio is 49.10\%. It then increases to 74.88\% after mitigation, quantifying a 52.51\% improvement as a result of bias mitigation.

\subsubsection{Equalized Odds Difference}
Figure~\ref{equalizedoddsdifference} depicts the equalized odds differences before and after bias mitigation. We find that the equalized odds difference is 17.70\% before mitigation and 8.10\% after mitigation. Therefore, mitigation helps to improve the equalized odds difference metric by 54.26\%.

\subsubsection{Statistical Significance Test}
Similar to the previous case, we use a \emph{Student's Two-sample t-test} to understand if the difference in the average equalized odds differences before and after mitigation is statistically significant. In the \emph{Student's Two-sample t-test}, the null hypothesis is \emph{$H_0: \mu_b = \mu_a$}. Here, \emph{$\mu_b$} is the average equalized odds difference before mitigation, and \emph{$\mu_a$} is the average equalized odds difference after mitigation. Our results reveal \emph{$t(58) = 6.43$} and \emph{p-value = $2.66e^{-8}$}, allowing the rejection of the null hypothesis since the \emph{p-value} is less than \emph{$\alpha = 0.05$}. Therefore, we can state that there is a significant difference between the average equalized odds differences before mitigation and after mitigation.

\section{Discussion}\label{discussion}

To our best knowledge, this is the first work that tries to mitigate the bias present in a stress/fatigue dataset collected from pilot and non-pilot college students and the machine learning models trained with the biased dataset. We utilized two bias mitigation algorithms, i.e., threshold optimizer with the demographic party and equalized odds constraints, to mitigate the sex, i.e., female and male, bias. Through detailed analysis, we find that our mitigation algorithms significantly reduce bias in both cases. 

This work has some limitations. First, our experimental findings are based on a limited subject pool. However, we perform cross-validation with random instance selection. Therefore, the findings still show some promise in this direction. Another limitation of this work is that we focused only on sex. However, other sensitive attributes, such as age and ethnicity, can also affect the performance of a model, and they have a higher impact on model performance. However, this is beyond the scope of this work. To better understand other types of sensitive attributes and before generalizing our findings, we need to conduct a large-scale longitudinal study with a bigger pool of diverse subjects with varying demographics and professional backgrounds. 

Another important direction is to investigate the effect of bias and its mitigation on objective sensor data-driven machine learning models that will be used to continuously detect stress and fatigue through various built-in sensors in smartphones, smartwatches, and the environment, e.g., cockpit. This bias mitigation work can further be applied to other areas, such as biometric authentication~\cite{dibbo2022phone} and other health monitoring~\cite{chen2020estimating,vhaduri2023transfer}, where bias analysis was not present.

\bibliographystyle{IEEEbib}
\bibliography{main}

\end{document}